\documentclass[runningheads]{llncs}
\usepackage[T1]{fontenc}
\usepackage{graphicx}
\usepackage{booktabs}
\usepackage[misc]{ifsym}
\newcommand{\corr}{(\Letter)}
\usepackage{tabularx}
\usepackage{hyperref}
\usepackage{comment}
\usepackage{xurl}
\begin{document}

\title{Can we Evaluate RAGs with Synthetic Data?}
\titlerunning{Can we Evaluate RAGs with Synthetic Data?}




\author{Jonas van Elburg \corr \inst{1,2}\orcidID{0009-0009-6917-8679} \and Peter van der Putten \inst{2,3}\orcidID{0000-0002-6507-6896}  
\and Maarten Marx \inst{1}\orcidID{0000-0003-3255-3729}}

\authorrunning{J. van Elburg et al.}


\institute{
IRLab, Informatics Institute, UvA, Amsterdam, the Netherlands 
\and AI Lab, Pegasystems, Amsterdam, the Netherlands
\and LIACS, Leiden University, Leiden, the Netherlands
}

\maketitle              

\begin{abstract}
We investigate whether synthetic question-answer (QA) data generated by large language models (LLMs) can serve as an effective proxy for human-labeled benchmarks when the latter is unavailable. We assess the reliability of synthetic benchmarks across two experiments: one varying retriever parameters while keeping the generator fixed, and another varying the generator with fixed retriever parameters. Across four datasets, of which two open-domain and two proprietary, we find that synthetic benchmarks reliably rank the RAGs varying in terms of retriever configuration, aligning well with human-labeled benchmark baselines. However, they do not consistently produce reliable RAG rankings when comparing generator architectures. The breakdown possibly arises from a combination of task mismatch between the synthetic and human benchmarks, and stylistic bias favoring certain generators. 

\end{abstract}

\section{Introduction}
Retrieval-Augmented Generation (RAG) has been proposed as a method for mitigating inaccurate and contextually inappropriate responses in domain-specific applications of Large Language Models (LLMs) \cite{ji2023survey,rawte2023survey,maynez-etal-2020-faithfulness}.  By incorporating factual information from external documents into the prompt, RAG models enhance explainability and domain specificity whilst reducing hallucinations \cite{lewis2020retrieval,li2024enhancing,huang2023survey,gao2023retrieval}. Commercial products such as Pega Knowledge Buddy (KB), on which this study is based, leverage this technology to facilitate a simple way of creating domain-specific question answering (QA) systems suited to a wide range of internal and external use cases, from human resource management, customer service, sales, to customer due diligence.  Since the relationship between RAG configuration and its downstream performance is complex, RAGs require rigorous benchmarking, especially in regulatory, cost- and error-sensitive environments \cite{agrawal2024mindful,barnett2024seven}.

Most RAG benchmarks rely on large open-domain datasets \cite{saad2023ares,rajpurkar2016squad,bajaj2016ms,petroni2020kilt,yu2024reeval,yu2024evaluation,stelmakh2022asqa}, but these may not accurately represent the domain-specific nature of typical RAG applications. Some evaluations leverage manually labeled domain-specific datasets \cite{xiong2024benchmarking,xu2024let,yu2024reeval}, but this approach is costly and inefficient, severely limiting the swift roll-out and adoption of new RAGs.

Instead, selected domain-specific research has opted for synthetic QA datasets \cite{es2023ragas,liu2023recall,chen2024benchmarking,tang2024multihop,wang2024domainrag,purwar2024evaluating,brehme2025llmstrustedevaluatingrag}. Generating synthetic QA pairs based on domain knowledge enables the construction and reconstruction of an up-to-date domain-specific dataset with pseudo-ground-truth answers. Although synthetic benchmarks pose an attractive solution for the automated evaluation of domain-specific RAG models, their reliability has not been systematically assessed in the context of design choices made with real benchmark data, raising concerns about the ecological validity of the method. To address this gap, we structure our study around the following two research questions:  how well do synthetic QA benchmarks align with human-annotated benchmarks in ranking question-answering RAG systems on quality, and can synthetic benchmarks serve as an effective method for automated evaluation of domain-specific RAG models.

To address these questions, we systematically assess the impact of the dataset generation method on RAG evaluation in both general and domain-specific settings. The presented experiments evaluate the consistency between rankings of several RAG systems using human-annotated and synthetic benchmarks. The results suggest that our synthetic benchmarks are suitable for retriever parameter tuning (e.g., number of context documents), but their reliability for evaluating generator parameters such as model choice is more limited.

The remainder of this paper is structured as follows: Section \ref{sec:rw} briefly reviews the most relevant related work. Sections \ref{sec:methods} and \ref{sec:experiments} explain our methodology and experimental setup for assessing synthetic benchmark reliability by comparing model rankings from synthetic to human-annotated benchmarks. Our results across two core experiments are presented in section \ref{sec:results}: Experiment A evaluates the consistency of synthetic and human benchmarks in ranking RAG models with varying retrieval parameters, and experiment B repeats the experiment ranking RAG models with varying pre-trained LLM architectures. Section \ref{sec:discussion} interprets the experiments in the context of our research questions, and section \ref{sec:conclusion} outlines the main conclusions.

\section{Related work}
\label{sec:rw}
The most relevant benchmarks relying on synthetic QA data include \textit{RGB} \cite{chen2024benchmarking}, \textit{Ragas} \cite{es2023ragas} and \textit{ARES} \cite{saad2023ares}. In all three benchmarks, QA pairs are generated by prompting an LLM to generate the questions and reference answers based on specific chunks of the corpus. \textit{ARES} further filters these data based on the round-trip consistency criterion, meaning only questions that are successfully mapped to their source document chunk by a semantic search were included in the evaluation dataset.

\textit{RGB} relies on classical supervised metrics such as exact match to rank models. In comparison, \textit{Ragas} and \textit{ARES} both include LLM-as-a-judge algorithms to evaluate answers without the need for a reference answer. It has been noted that these evaluation algorithms can be sensitive to the prompt design \cite{saad2023ares}. To mitigate this, \textit{ARES} calibrates the judges using 150 binary labeled human-preference data-points per domain, and further provides domain-specific few-shot examples in the prompt.

ARES is the only benchmark that explicitly evaluates ranking consistency with ground-truth rankings, demonstrating high agreement across datasets \cite{saad2023ares}. However, because it relies on a small set of human-labeled validation data to calibrate its judges, its performance may not generalize when using only synthetic data, different evaluation metrics, or in highly domain-specific settings.

\section{Methods}
\label{sec:methods}

RAG architectures operate by chunking corpus documents into overlapping fixed-size segments and storing these in a knowledge base. At query time, a dense retriever identifies the top-$k$ semantically relevant chunks, subject to a semantic similarity threshold, which are then injected into a prompt template that is fed to the LLM for answer generation \cite{lewis2020retrieval}.

To assess the effectiveness of generated QA pairs in benchmarking domain-specific RAG systems, we compare the rankings of a set of RAGs derived from two benchmarks. Our \textit{Human Benchmark} uses a dataset of manually curated questions and annotated answers, while the \textit{Synthetic Benchmark} only uses generated QA pairs.

The RAGs are evaluated on a standard set of evaluation metrics described in section \ref{sec:scoring_metrics}, and ranked according to their performance across these metrics. Finally, we compare rankings from the synthetic benchmark to those from the human benchmark, treating the latter as a gold-standard reference. The consistency between the rankings thus produces a measure of reliability of the synthetic benchmark.

\section{Experiments}
\label{sec:experiments}
This section describes the experimental conditions under which we test the validity of synthetic benchmarks for RAG systems. We detail the default individual model configuration and evaluation, datasets used, the synthetic data generation process, model variations evaluated, and our benchmark evaluation procedure.\footnote{Code and supplementary materials are available at \url{https://github.com/JonasElburgUVA/Can-we-evaluate-RAGs-with-synthetic-data/blob/master}.}

\subsection{RAG system set-up}
RAGs were implemented on the KB platform\footnote{See \url{https://www.pega.com/products/genai-knowledge-buddy}.}, but can be reproduced outside of this environment. The RAGs employ a dense retriever, which searches the knowledge base using cosine similarity of semantic OpenAI ada-002 embeddings. The template prompt to which the retrieved documents are added is shown in the \href{https://github.com/JonasElburgUVA/Can-we-evaluate-RAGs-with-synthetic-data/blob/master/SupplementaryMaterials.pdf}{supplementary materials (Listing 1)}. All hyperparameters of our baseline model are presented in Table \ref{tab:BaseHyperparameters}.  

\begin{table}[tbp]
    \centering
    \caption{Default hyper-parameters of the baseline RAG.}
    \label{tab:BaseHyperparameters}
    \begin{tabular}{ll} 
        \toprule
        \textbf{Chunk size (words)}& 1000 \\ 
        \textbf{Chunk overlap (words)}& 200 \\ 
        \textbf{Min retrieval similarity score} & 80.00 \\ 
        \textbf{Max retrieved documents} $k$& 5 \\ 
        \textbf{Temperature} & 0.0 \\
 \textbf{Top-P}&0.95\\ 
        \textbf{Generator architecture} & gpt-4o-mini\\ 
        \textbf{Embedding model} & text-embeddings-ada-002 (version 2)\\ 
        \bottomrule
    \end{tabular}

\end{table}

\subsection{RAG scoring}
\label{sec:scoring_metrics}
The resulting RAG systems are evaluated in terms of their answer accuracy compared to a reference answer using the \textit{Ragas} framework \cite{es2023ragas}. We assess answer accuracy using the key-word based ROUGE-L F1, BLEU, and Levenshtein string similarity, alongside semantic similarity measured with OpenAI text-embedding-3-small. Additionally, three LLM-based evaluation metrics from the \textit{Ragas} framework are implemented: Answer relevance, Faithfulness, and Context Precision (LLM-based MAP@k), all using GPT-4o-mini and the prompts provided by \textit{Ragas} to assess different aspects of the given answer \cite{es2023ragas}. More details on these metrics can be found in the \href{https://github.com/JonasElburgUVA/Can-we-evaluate-RAGs-with-synthetic-data/blob/master/SupplementaryMaterials.pdf}{supplementary materials}.

The template prompt instructs the RAG models to reject questions when deemed unanswerable based on retrieved context. Since only answerable questions are considered in this study, we further compute the false negative (FN) rate defined as the rate of question rejections. All other metrics are set to zero for rejected responses.

\subsection{Datasets}
\label{sec:methods_data}
This work includes four datasets; two commonly used open-domain datasets, and two novel domain-specific datasets.

\emph{SQuAD} \cite{rajpurkar2016squad} is an open-domain dataset constructed to evaluate extractive question-answering models. The corpus is built from Wikipedia pages, with 107785 questions generated via crowd-sourcing and answers extracted directly from the text.

\emph{ASQA} \cite{stelmakh2022asqa} is an open-domain long-form question-answering dataset containing ambiguous questions and human-annotated answers disambiguating the possible responses based on Wikipedia articles. The 5301 publicly available questions do not all come with long-form annotated answers, and not all annotations cite (existing) web pages. We therefore included a few pre-processing steps, limiting the dataset to questions with at least one long-form answer annotation which cites at least one Wikipedia page available at the time of web scraping (March 6, 2025).

\emph{Launchpad} is a novel domain-specific QA dataset based on the documentation of the Pega product \textit{Launchpad} \cite{Pega}. Questions were obtained by retrieving real queries asked to an internally deployed \textit{Launchpad Buddy}; a RAG model conditioned on the launchpad documentation. The queries were filtered by a domain expert to only include relevant and sensible questions. The model's generated answers were further labeled and edited by the expert to construct a dataset of 94 validated QA pairs.

\emph{Sales} just like Launchpad, is a dataset sourced internally consisting of a corpus of sales-related documentation only partially available on the public internet, and 99 real questions with edited responses sourced from conversations with the \textit{Sales Buddy}.

To manage computational resources, we limit the open-domain benchmarks to 100 randomly sampled questions. For SQuAD, we still use the full Wikipedia corpus, allowing the retriever to access all original documents. In contrast, for ASQA, we restrict the corpus to only the 169 Wikipedia pages cited in the selected reference answers to avoid unnecessary web scraping. 

\subsection{Synthetic QA generation}
To generate synthetic QA pairs based on both datasets, we adapt prompts described by LlamaIndex \cite{Theja_2023}, requesting questions and answers that must be answerable given a document from our corpora (\href{https://github.com/JonasElburgUVA/Can-we-evaluate-RAGs-with-synthetic-data/blob/master/SupplementaryMaterials.pdf}{Supplementary materials Listing 2}). The LLM is GPT-4o version 2024-11-20 with temperature set to 0.7 and top-$p$ set to 0.95. We exclude self-validating techniques such as round-trip consistency to obtain an unbiased baseline performance of naive synthetic data generation. 

We sample 100 synthetic QA pairs per dataset using a default of 20 randomly sampled documents, meaning we obtain five questions per document. For the Launchpad dataset this is adjusted to two questions per document and 50 sampled documents due to a lack of lengthy documents in the corpus (Table \ref{tab:questions_descriptives}). For the Sales dataset, which has a similarly low average document length but far more total available documents, we instead employ a minimum document length of 500 words.

\begin{table}[tbp]
\centering
\caption{Descriptive statistics of the real and synthetic questions and reference answers.}
\label{tab:questions_descriptives}
\begin{tabular}{l|cccc}
\toprule
 & \textbf{SQuAD} & \textbf{ASQA} & \textbf{Launchpad}& \textbf{Sales} \\
\midrule
Total Documents & 475 & 166 & 378 & 29987 \\
Sampled Documents & 20 & 20 & 50& 20 \\
Words per Document ($\mu \pm \sigma$)& 4319 $\pm$ 1678 & 3522 $\pm$ 4284 & 488 $\pm$ 405 & 645 $\pm$ 1146 \\
Human Questions & 100 & 100 & 94 & 99 \\
Synthetic Questions & 100 & 100 & 100 & 100 \\
\bottomrule
\end{tabular}
\end{table}

\subsection{Experimental Design}
To evaluate the validity of synthetic benchmarks, we require a set of RAG systems to be evaluated and ranked. We created alternative configurations across two dimensions in two separate experiments.

\emph{Experiment A} compares the baseline RAG configuration (Table~\ref{tab:BaseHyperparameters}) with three alternative systems that each modify the retrieval behavior. In the first variant, the retriever is limited to returning only one chunk, reduced from the baseline setting of five. The second variant increases this limit to 10 chunks. The third variant also retrieves 10 chunks but disables the minimum similarity threshold, returning the top 10 results regardless of how weakly they match the query.

\emph{Experiment B} evaluates the baseline model next to four alternative models with different generator model choices. The alternative generators employed are GPT-3.5, GPT-4o, Llama-7b-instruct, and Claude-3-Haiku. Results for Llama-7b-instruct on the sales dataset are excluded due to the model’s deprecation within the Pega infrastructure during the course of this study.

\subsection{Evaluation}

Given a set of questions, retrieved contexts, generated answers and reference answers we calculate the evaluation metrics described in section \ref{sec:scoring_metrics} for each model. The evaluation metrics are used to construct rankings of the RAG systems according to the human and synthetic benchmarks. We then calculate the Kendall rank correlation between the two rankings for each metric of interest, providing a statistic reflecting how well the synthetic benchmark matches the human benchmark under each metric. A perfectly consistent ranking produces $\tau=1.0$, whereas a completely inverted ranking produces $\tau=-1.0$. The statistic is not evaluated for statistical significance due to the limited number of models evaluated.

\section{Results}
\label{sec:results}

In this section we discuss our findings experiment by experiment. We begin by detailing the results of the data acquisition process. Subsequently, we discuss the results of experiment A, assessing the efficacy of synthetic benchmarks in ranking models with differing retriever parameters. Finally we present the results of experiment B, analyzing whether we can produce consistent rankings of the generating LLM model used.

\subsection{Data}

\begin{figure}[tbp]
    \centering
    \includegraphics[width=0.7\linewidth]{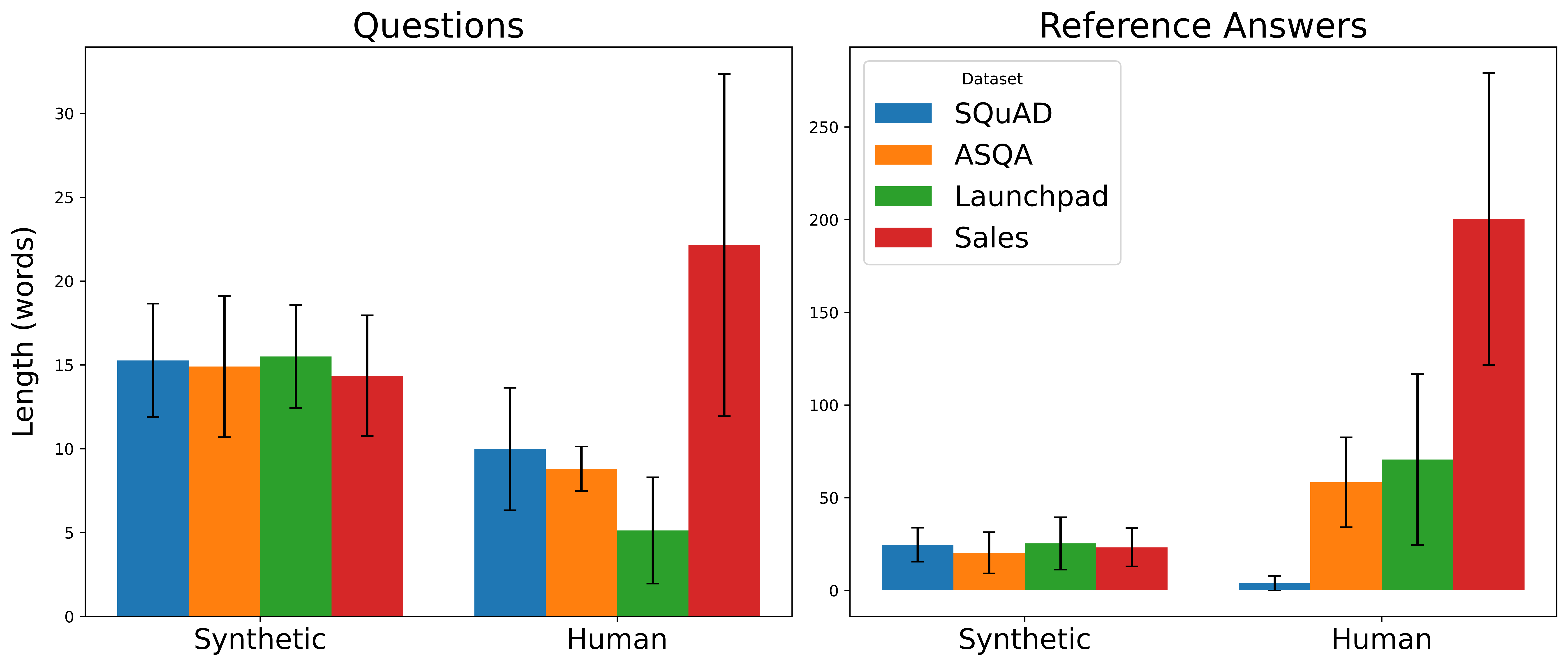}
    \caption{The average length of synthetic and human questions and reference answers of all datasets. Error bars indicate standard deviations.}
    \label{fig:qa-descriptives}
\end{figure}

The four datasets differ in question format. QA pairs in the SQuAD dataset follow an extractive nature, whereas questions in the other datasets are open-ended, often requiring generative reasoning (Table \ref{tab:example-questions-human}). This distinction is reflected in reference answer lengths: SQuAD answers are short (3.96 $\pm$ 5.22 words), whereas Launchpad (70 ± 46.12 words) and Sales (204 $\pm$ 78.83) answers are much longer. We further observe that the questions in the Sales benchmark are longer than other human benchmarks. These differences are not observed between the synthetic datasets (Figure \ref{fig:qa-descriptives}).

The example QA pairs in Tables \ref{tab:example-questions-human}-\ref{tab:example-questoins-gen} highlight key differences between the human and synthetic benchmarks. Across ASQA, Launchpad, and Sales, synthetic questions are generally more specific and technical. This is especially evident in ASQA, where human-written questions are designed to be ambiguous, while ASQA-Gen features more focused queries (e.g. asking about a specific award). Human questions, drawn from real users, are often more general or even vague.

An exception is SQuAD, where the human benchmark contains very specific extractive questions leading to a different kind of task mismatch: extractive queries in the human benchmark compared to abstractive queries in the synthetic benchmark. These differences highlight an important source of task mismatch between the two benchmark types.

\begin{table}[tbp]
    \centering
    \caption{Example question-answer pairs from regular datasets.}
    \label{tab:example-questions-human}
    
    \begin{tabularx}{\linewidth}{|X| >{\hsize=0.3\textwidth}X| >{\hsize=0.52\textwidth}X|}
        \toprule
         Dataset & Question & Reference Answer \\
         \midrule
         SQuAD & What is the original meaning of the word Norman? & Viking \\
         \midrule
         ASQA & How many state parks are there in Virginia? & When the Virginia state park system was formed on June 15, 1936, there were only six state parks in the entire state. As of 2016, that number had gone up to 38 state parks. \\
         \midrule
         Launchpad & What is a query in data connection? & When configuring a data connection, you can select a query to specify how information is fetched \\
         \midrule
         Sales & Can Pega GenAI coach be used on a mobile device? & Yes, Pega GenAI Coach can be used on a mobile device. The Pega Sales Automation mobile application includes support for Pega GenAI Coach, which allows sales representatives to generate personalized summaries and instructions directly from their mobile devices. This feature is designed to help users evaluate and summarize opportunity data, providing recommendations and instructions to successfully close opportunities. \\ 
         \bottomrule
    \end{tabularx}
    
\end{table}

\begin{table}[btp]
    \centering
    \caption{Example question-answer pairs from synthetic datasets.}
    \label{tab:example-questoins-gen}
    
    \begin{tabularx}{\linewidth}{|X| >{\hsize=0.3\textwidth}X| >{\hsize=0.52\textwidth}X|}
        \toprule
         Dataset & Synthetic question & Synthetic reference answer \\
         \midrule
         SQuAD & What was the origin of the Normans and how did they come to settle in Normandy? & The Normans were descended from Norse raiders and pirates from Denmark, Iceland, and Norway who settled in Normandy after swearing fealty to King Charles III of West Francia.\\ 
         \midrule
         ASQA & What award did Ethan Peck win at the 2009 Sonoma International Film Festival? & He won the award for 'Best Actor' for his portrayal of 'Sailor'. \\
         \midrule
         Launchpad & What is the primary benefit of adding built-on Rulesets in application development? & Save time and reduce development costs by reusing elements between your applications. \\
         \midrule
         Sales & How can businesses adjust text analysis to meet their specific needs? & By configuring advanced settings, you can adjust text analysis to your business-specific needs. \\
         \bottomrule
    \end{tabularx}
    
\end{table}

\subsection{Experiment A - Retrieval}

The first experiment evaluates how the amount of retrieved chunks and the presence of a minimum ranking score threshold affect the ranking of RAG systems across human and synthetic benchmarks. This setup isolates the impact of retrieval quality on answer generation, keeping the generator model constant.

\begin{figure}[tbp]
  \centering
  \mbox{}\hfill  
  \begin{minipage}[b]{0.475\textwidth}
    \centering
    \includegraphics[width=\textwidth]{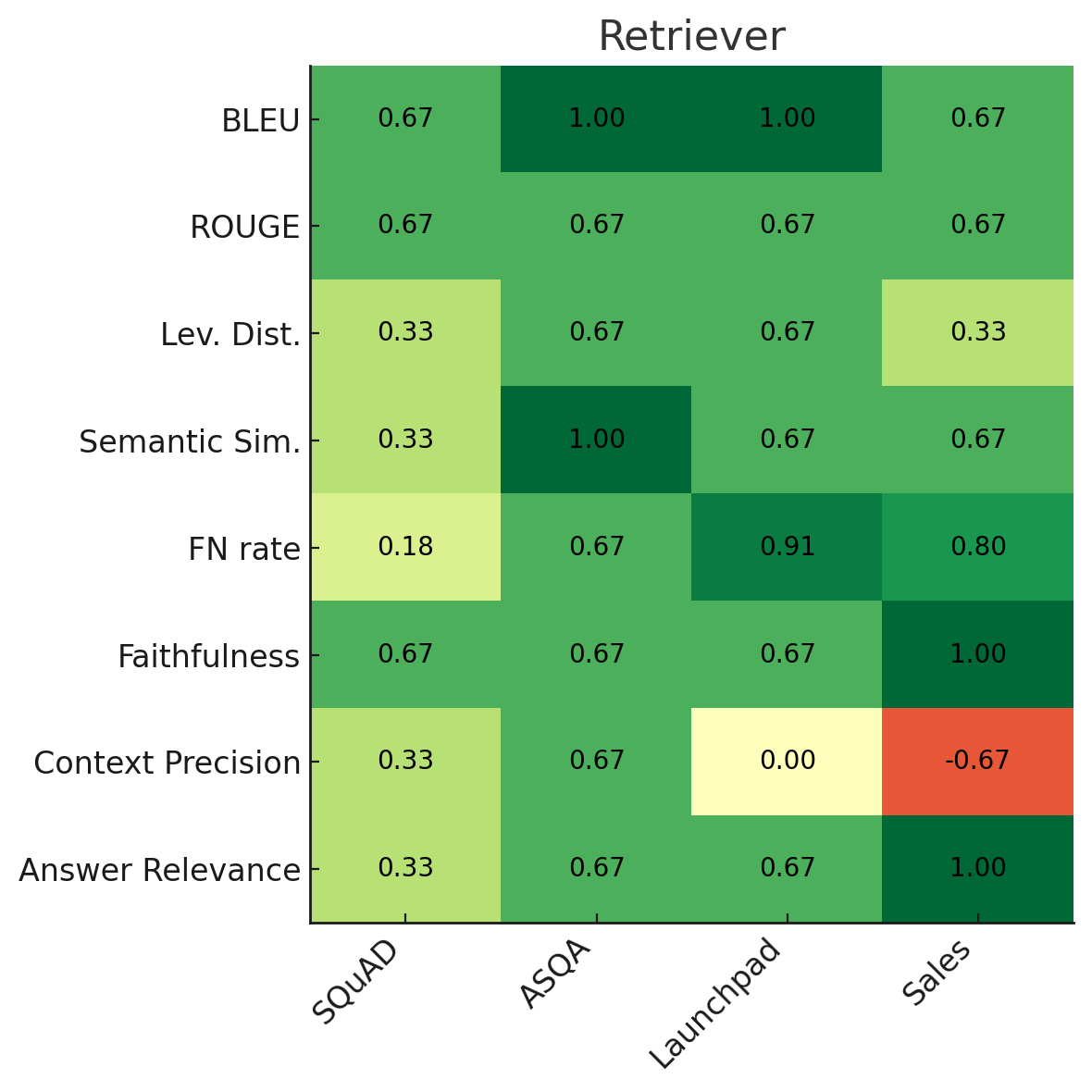}
  \end{minipage}
  \hfill
  \begin{minipage}[b]{0.5\textwidth}
    \centering
    \includegraphics[width=\textwidth]{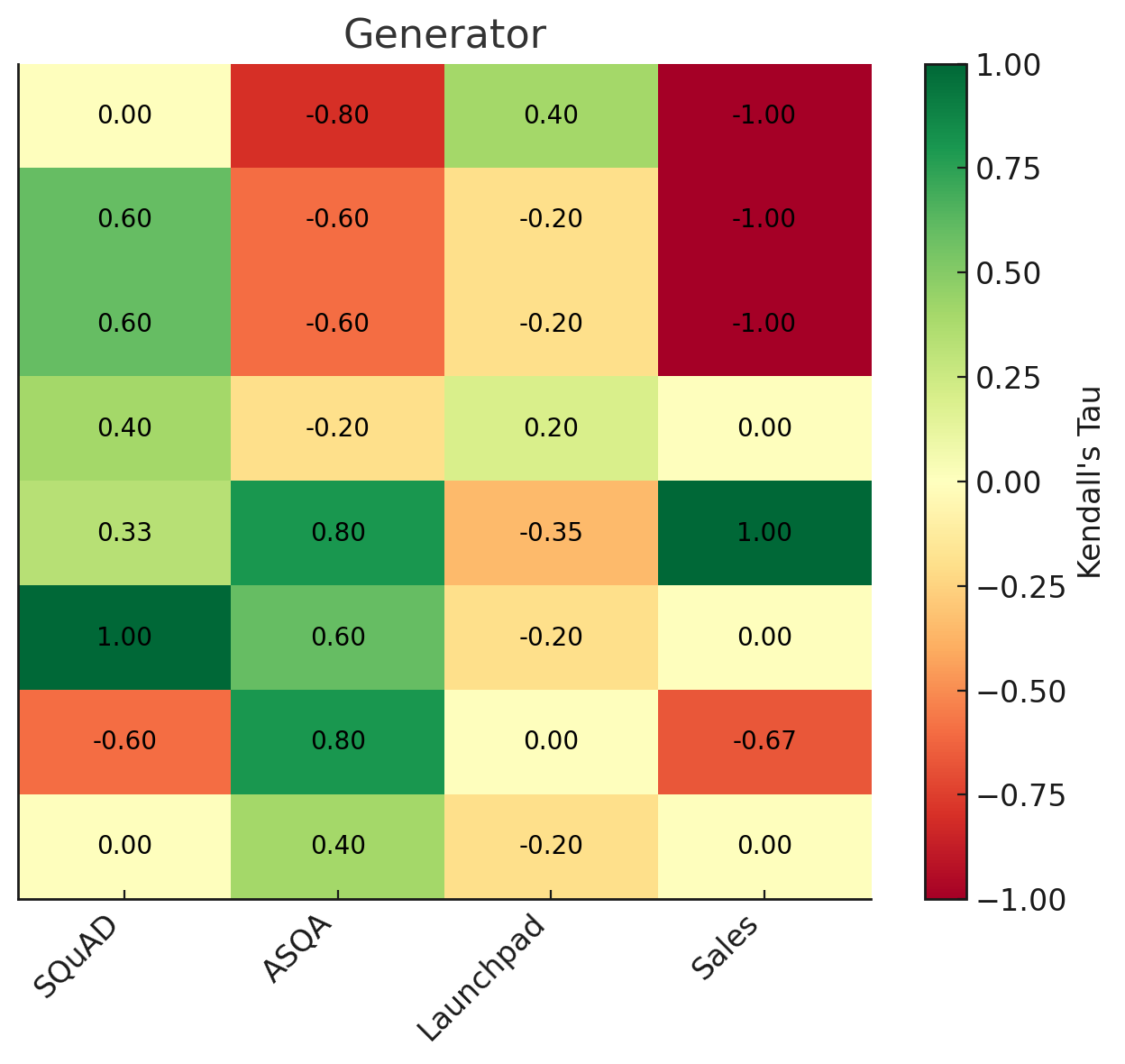}
  \end{minipage}
  \mbox{}\hfill
  \caption{Kendall’s $\tau$ rank correlation coefficients between human and synthetic benchmarks across four datasets and two experiments. Rank correlation is generally higher in the retriever experiment than the generator experiment. Furthermore, consistency is to an extent dataset- and metric-dependent.}
  \label{fig:kendall-both}
\end{figure}

We observe moderate to high ranking consistency between human and synthetic benchmarks across all datasets and evaluation metrics, with the SQuAD dataset resulting in slightly lower scores than other datasets (Figure \ref{fig:kendall-both}). The difference between synthetic and human QA length observed in Figure \ref{fig:qa-descriptives} indicates that SQuAD and Sales may have suffered from a higher format mismatch than ASQA and Launchpad. These results suggest that ranking agreement between synthetic and human benchmarks is to an extent dataset-dependent.  

\begin{figure}[tbp]
    \centering
    \includegraphics[width=0.7\linewidth]{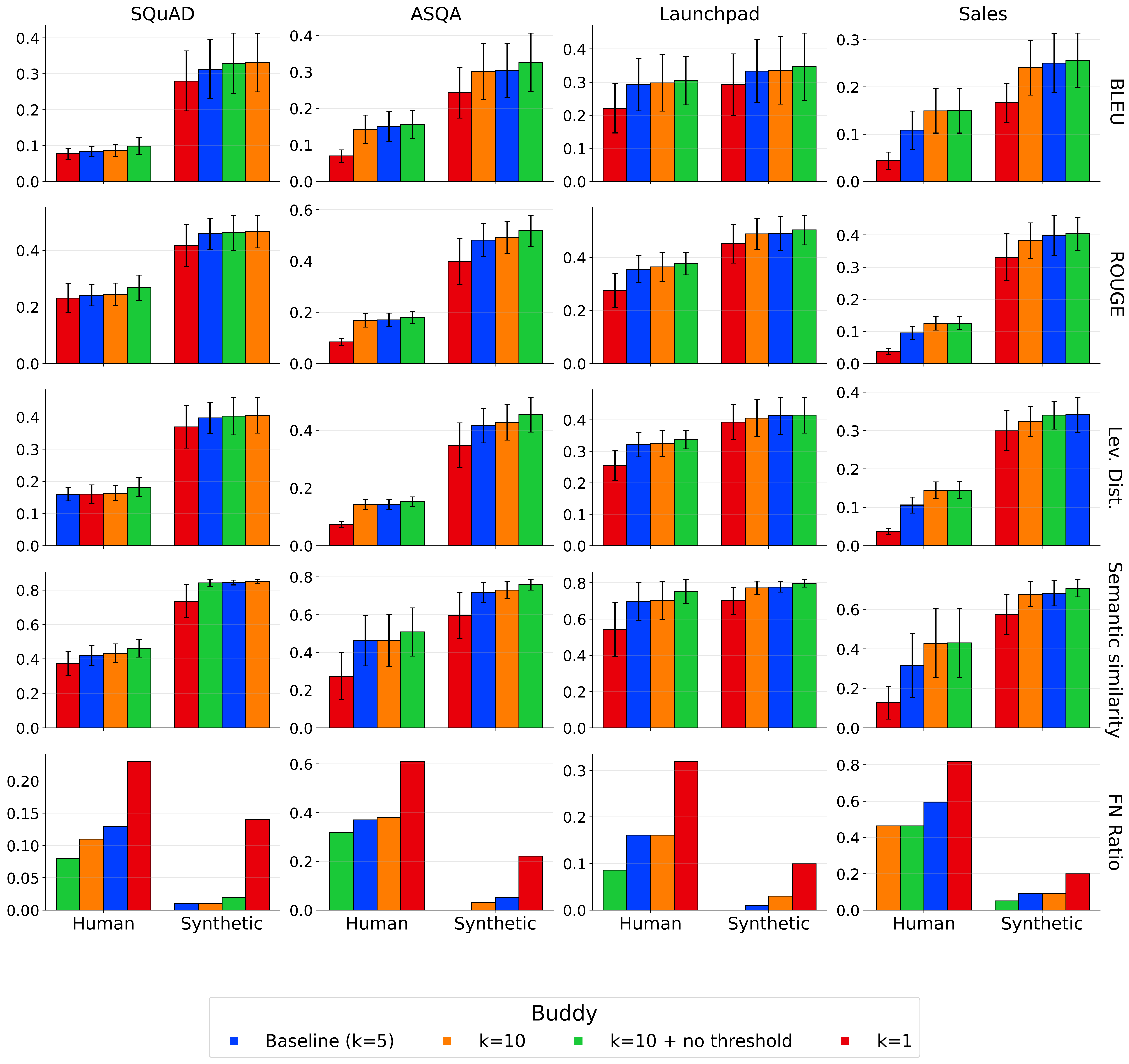}
    \caption{Supervised metrics in the retrieval experiment. Error bars represent the variance over the 94-100 datapoints.}
    \label{fig:retrieval-supervised-vis}
\end{figure}

\begin{figure}[tbp]
    \centering
    \includegraphics[width=0.7\linewidth]{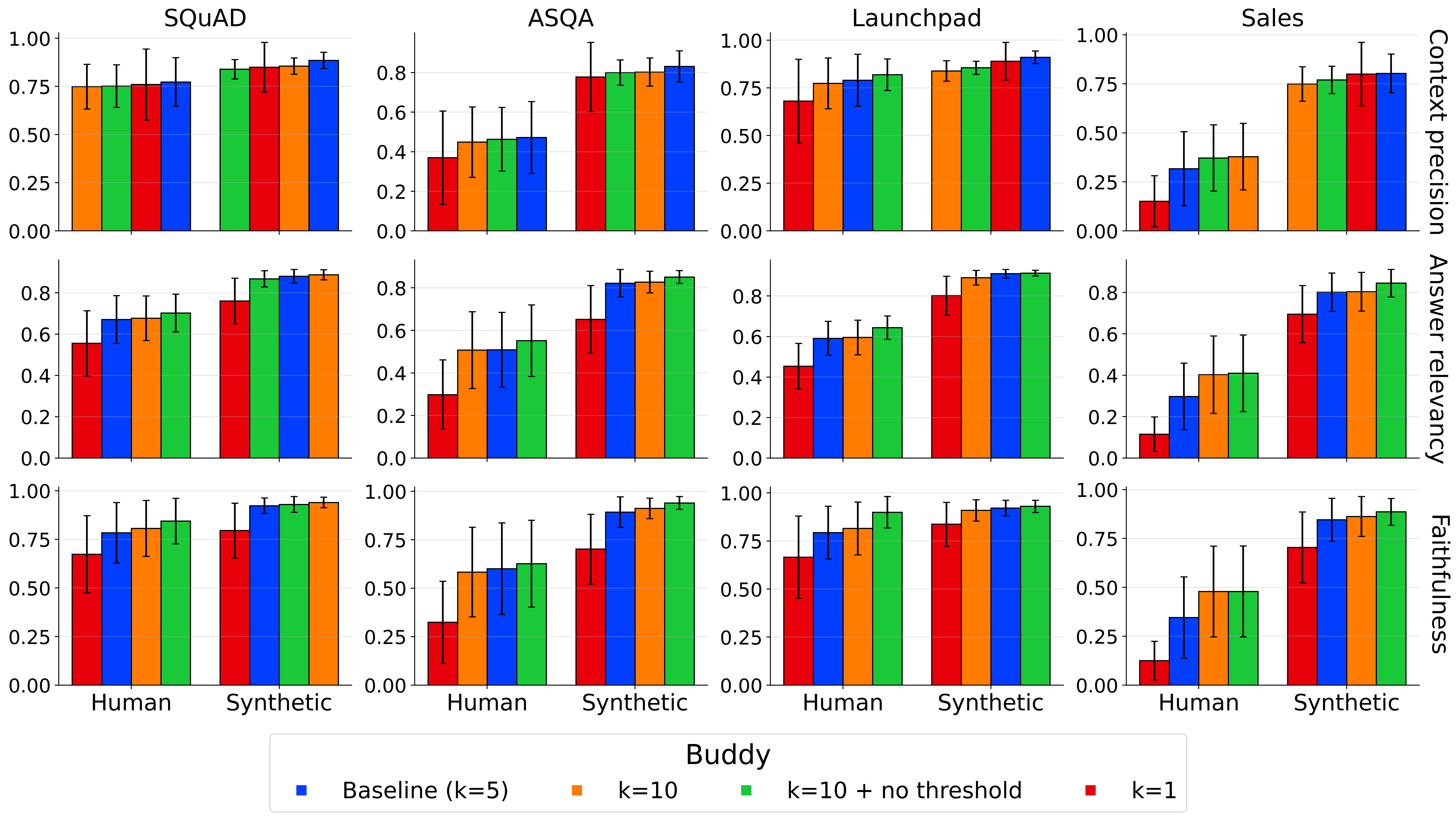}
    \caption{LLM-based metrics in the retrieval experiment. Error-bars represent the variance over the 94-100 datapoints.}
    \label{fig:retrieval-llm-vis}
    \end{figure}

Across most datasets and metrics, RAG systems with more retrieved context are preferred to those with a lower $k$ (Figures \ref{fig:retrieval-supervised-vis}–\ref{fig:retrieval-llm-vis}). Furthermore, RAGs consistently perform worse on human benchmarks than on their synthetic counterparts, indicating an underestimated task difficulty in the synthetic benchmarks.

Considering the different metrics used, we find the highest average ranking consistency using the keyword-based and precision-oriented BLEU metric ($\tau_{\mu}=0.84$). ROUGE (keyword-based, recall-oriented) and Semantic Similarity (embedding-based) also produce strong alignment ($\tau_{\mu}=0.67$), suggesting that both surface-level and semantic supervised measures of correctness can yield consistent rankings.

The unsupervised Faithfulness and Answer Relevance metrics produce consistent rankings across datasets, and prefer higher context-windows, similar to more standard supervised metrics such as semantic similarity. In contrast, Context Precision reveals notable discrepancies between human and synthetic benchmarks ($\tau_{\mu}=0.08$). In synthetic datasets, Context Precision scores are high across all retrieval settings, indicating the most relevant chunk is often ranked highest by the semantic search and is sufficient for answering the question (Figure \ref{fig:retrieval-llm-vis}). In contrast, most human benchmarks show a clear performance gap: the low-context model consistently underperforms, indicating that single top-ranked documents are often insufficient for answering questions. This again suggests that the synthetic benchmark may underestimate the retrieval challenge posed by natural user queries.

\subsection{Experiment B - LLMs}

\begin{figure}[tbp]
    \centering
    \includegraphics[width=0.7\linewidth]{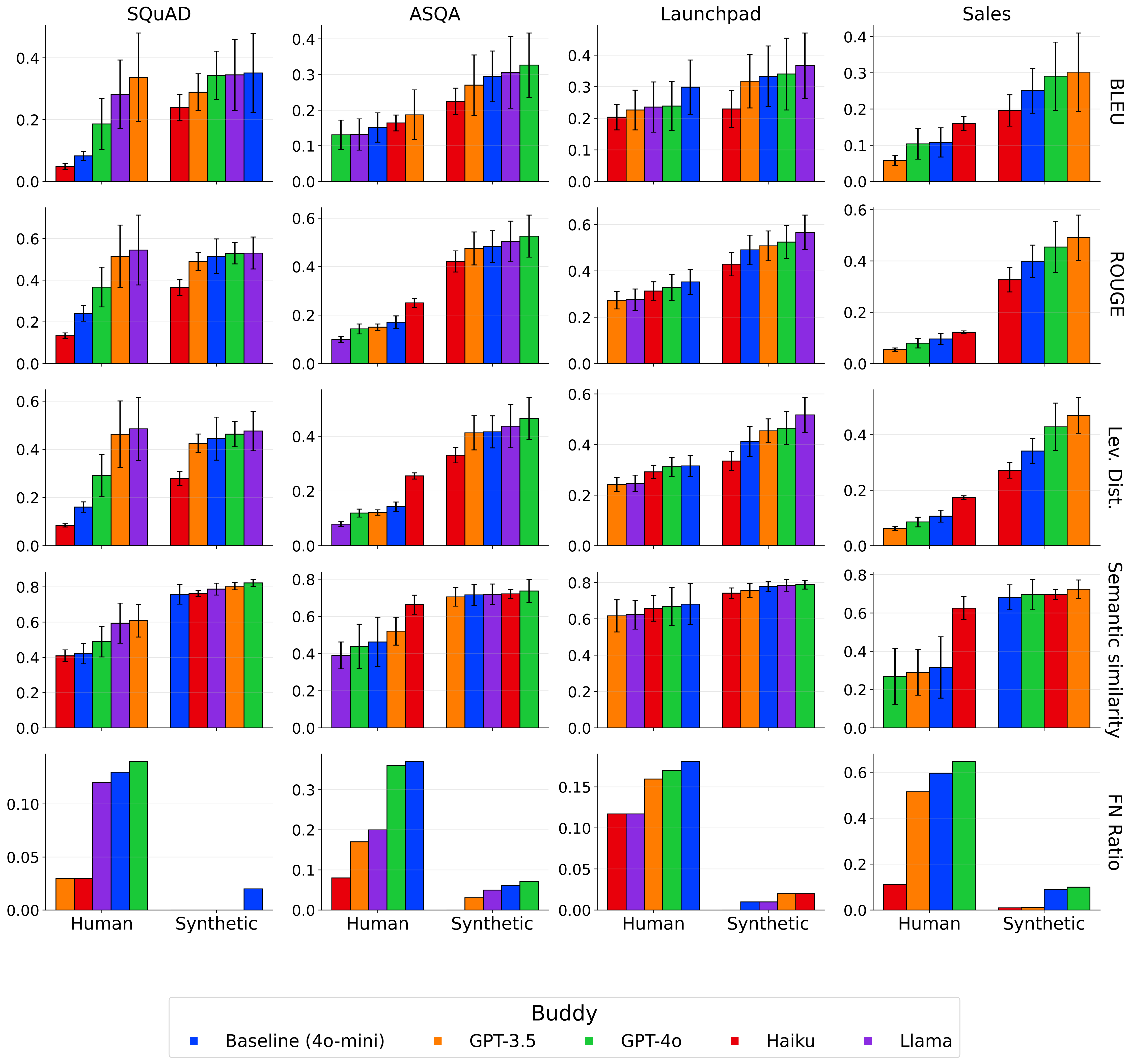}
    \caption{Supervised metrics in the LLM choice experiment. Error bars represent the variance over the 94-100 datapoints. The Llama model is missing from the Sales experiment since it was removed from the KB platform since the start of the research project.}
    \label{fig:models-supervised-vis}
\end{figure}

\begin{figure}[tbp]
    \centering
    \includegraphics[width=0.7\linewidth]{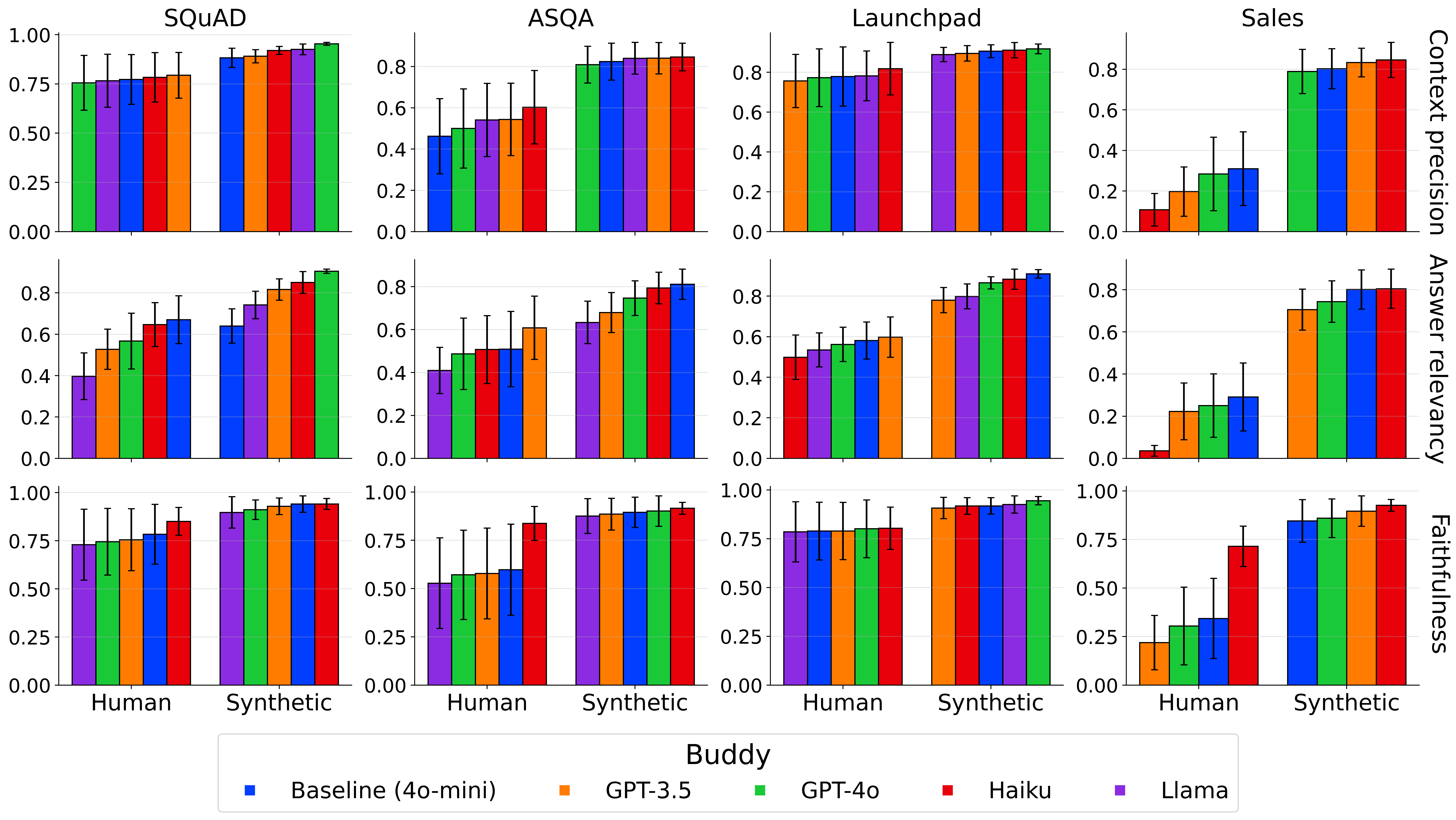}
    \caption{Unsupervised metrics in the LLM choice experiment. Error bars represent the variance over the 94-100 datapoints. The Llama model is missing from the Sales experiment since it was removed from the KB platform since the start of the research project.}
    \label{fig:models-llm-vis}
\end{figure}

Experiment B evaluates the consistency of model rankings across five large language models (LLMs) as part of RAG systems using both human and synthetic evaluation benchmarks. Retrieval settings are held constant to isolate the effect of the generator model choice (Figure~\ref{fig:models-supervised-vis} and \ref{fig:models-llm-vis}).

Overall, rankings differ substantially between the synthetic and human benchmarks across datasets and evaluation metrics (Figure \ref{fig:kendall-both}). Several metric–dataset combinations yield low or even negative Kendall’s $\tau$ values, indicating little agreement and, in some cases, invert model preferences. These inconsistencies are observed across both supervised metrics and unsupervised LLM-based metrics such as Faithfulness and Answer Relevance, suggesting that the ranking differences are not solely due to variation in reference answers.

\section{Discussion}
\label{sec:discussion}

This study addresses a key gap in the evaluation of synthetic QA benchmarks for RAG systems by systematically comparing model rankings under human and synthetic benchmarks across four datasets, including two highly domain-specific corpora. While prior evaluation benchmarks such as \textit{Ragas} and \textit{ARES} employ synthetic data, they either assume synthetic benchmark validity without comparison to human annotations or use additional labeled data. In contrast, we conduct two controlled experiments, one varying retriever parameters (Experiment A) and one comparing generator architectures (Experiment B), to assess ranking consistency under both classical supervised and LLM-based metrics. This setup enables a more fine-grained analysis of when synthetic benchmarks produce reliable rankings and highlights key factors that influence their alignment with human evaluation. The results especially have implications for the adaptation of domain-specific RAG evaluation frameworks, such as on the KB platform. The following sections reflect on these results and outline limitations to guide future research.

\subsection{Reflections on results}
Experiment A demonstrates moderate to high ranking consistency between human and synthetic benchmarks across datasets and evaluation metrics, with the exception of Context Precision. In both benchmark types, larger context windows generally correlates with higher scores, suggesting that additional input, even if only partially relevant, positively impacts model performance. The ranking consistency is highest for ASQA and Launchpad data, which likely contains less task mismatch originating from question and answer lengths than SQuAD and Sales (Figure \ref{fig:qa-descriptives}). These results suggest that synthetic benchmarks with minimized task mismatches can be used when optimizing retrieval parameters. 

In contrast to Experiment A, Experiment B shows substantial inconsistency in model rankings between human and synthetic benchmarks across datasets and metrics. While some metric–dataset combinations produce moderate to strong correlations, these cases are isolated, and no general trend of alignment emerged. 
A possible explanation for the lack of correlation is stylistic differences between the two benchmark types: synthetic questions are generally simpler and more specific, reducing task complexity and the diversity of valid answers. This likely impacts generator architectures more than retrieval settings. The varied output styles of different LLMs in Experiment B makes rankings sensitive to question formulation, especially in more open-ended and abstract questions. All models in Experiment A use the same LLM, producing stylistically similar outputs, which may have masked such effects. Moreover, since the synthetic data have been generated using GPT-4o, there may be a stylistic bias favoring this model in the synthetic benchmark, causing further ranking inconsistency.


The inconsistency in experiment B can  also be observed in unsupervised metrics, despite Faithfulness and Answer Relevance providing consistent results in experiment A. Since these metrics are independent of the reference answer, the task mismatch only present in the questions is sufficient to obfuscate which model was preferred under these metrics. 

\subsection{Known limitations}
This study faces limitations related to the datasets, QA generation, and evaluation methodology, largely due to technical constraints and project scope. Addressing these factors in future work can improve the robustness of synthetic benchmark evaluation.

The datasets used in this study pose several challenges. We observe a consistent performance gap between human and synthetic benchmarks across all methods, suggesting that synthetic benchmarks may underestimate task difficulty. This may be due to simpler questions and increased surface-level overlap between questions, corpus chunks and reference answers. The benchmarks also vary in question and reference answer length, especially visible in the SQuAD and Sales datasets, further contributing to task mismatches between synthetic and human benchmarks (Figure \ref{fig:qa-descriptives}).

The open-domain nature of both SQuAD and ASQA makes their results less applicable to the practical use case of many RAG applications, since the evaluated RAGs may have benefited from the presence of the source dataset in LLM pretraining. While the Launchpad and Sales datasets provide a more realistic domain-specific evaluation setting, we are unable to source more than 100 human-annotated QA pairs for these datasets. 

We purposely used only a single, simple synthetic QA generation method, without considering more complex synthetic QA pairs such as multi-hop or adversarial questions \cite{alber2025medical}. Furthermore, the prompt used to synthesize QA pairs is relatively simplistic and would likely benefit from custom few-shot examples, as was done in previous work \cite{torres2024automated,li2024enhancing,saad2023ares}. Additionally, we did not use self-validation methods such as round-trip consistency to filter the synthetic data, which may increase the reliability of the benchmark \cite{alberti2019synthetic,saad2023ares}. With additional calibration these factors could mitigate task mismatch and influence ranking consistency. 


\section{Conclusion}
\label{sec:conclusion}
We have introduced and evaluated a synthetic benchmarking pipeline for domain-specific RAG applications in the absence of ground-truth data. While previous work has already used similar pipelines  \cite{es2023ragas,liu2023recall,chen2024benchmarking,tang2024multihop,wang2024domainrag,purwar2024evaluating}, the impact of generating data on evaluation reliability has not been evaluated systematically prior to this study. 

Our findings suggest that synthetic benchmarks can provide a reliable signal when tuning retrieval parameters, particularly when the synthetic and human tasks are well-aligned in format and difficulty. However, when comparing different generator architectures, we observed substantial inconsistencies between synthetic and human benchmark rankings. These inconsistencies appear to stem from differences in question complexity and answer style, which had a disproportionate effect when model outputs varied more substantially. This highlights a key limitation in current synthetic evaluation pipelines: while scalable and efficient, they may under-represent the nuances and difficulty of real-world user queries. As such, synthetic benchmarks should not be treated as universally reliable, but rather as tools whose validity depends on the alignment between task design, metric choice, and evaluation target.

Future work should focus on increasing control over the difficulty and task-relevance of synthetic QA data, and attempt to extend the presented findings with additional datasets and models for more interpretable results.


\begin{credits}
\subsubsection{\ackname} We would like to thank Dawid Budynko for providing, labeling, and adjusting the Launchpad dataset, likewise Stephen Sanzo and Daniel Ford for the Sales dataset, Kishore Adiga for general data support, Stijn Kas for data science support and Andrei Adamian for making this project possible in the first place.

\subsubsection{\discintname} The authors have no competing interests to declare that are relevant to the content of this article.
\end{credits}
%
%
%
\bibliographystyle{splncs04}
\bibliography{references}
%






\end{document}